\documentclass[10pt,twocolumn,letterpaper]{article}

\usepackage{iccv}              

\definecolor{iccvblue}{rgb}{0.21,0.49,0.74}
\usepackage[pagebackref,breaklinks,colorlinks,allcolors=iccvblue]{hyperref}
\usepackage{amssymb}
\usepackage{amsmath, bm}
\usepackage[noend]{algpseudocode}
\usepackage{algorithm}
\usepackage{graphicx}
\usepackage{soul}
\usepackage{caption}
\usepackage{subcaption}
\usepackage{multirow}

\newcommand{\BlockTitle}[1]{\vspace{0.5em} \Statex \textbf{#1} \vspace{0.2em}}

\DeclareMathOperator*{\argmin}{arg\,min}

\newcommand{\bA}{\mathbf{A}}

\newcommand{\cD}{\mathcal{D}}

\newcommand{\bH}{\mathbf{H}}

\newcommand{\bc}{\mathbf{c}}

\newcommand{\bF}{\mathbf{F}}

\newcommand{\cS}{\mathcal{S}}

\newcommand{\bP}{\mathbf{P}}

\newcommand{\bp}{\mathbf{p}}
\newcommand{\bQ}{\mathbf{Q}}
\newcommand{\bq}{\mathbf{q}}

\newcommand{\bbR}{\mathbb{R}}

\newcommand{\bW}{\mathbf{W}}

\newcommand{\bX}{\mathbf{X}}
\newcommand{\bx}{\mathbf{x}}
\newcommand{\by}{\mathbf{y}}

\newcommand{\bz}{\mathbf{z}}

\newcommand{\btheta}{\boldsymbol\theta }

\newcommand{\rscpu}{RS-CPU}
\newcommand{\neurorscpu}{NeuroRF-CPU}

\newcommand{\neurorsloihi}{NeuroRF-Loihi}

 \def\paperID{14} 
\def\confName{ICCV}
\def\confYear{2025}

\title{Event-driven Robust Fitting on Neuromorphic Hardware}

\author{%
  Tam Ngoc-Bang Nguyen\textsuperscript{\textnormal{1}}, 
  Anh-Dzung Doan\textsuperscript{\textnormal{1}}, 
  Zhipeng Cai\textsuperscript{\textnormal{2}}, 
  Tat-Jun Chin\textsuperscript{\textnormal{1}}\\
  \textsuperscript{1} Australian Institute for Machine Learning, The University of Adelaide, \\
  \textsuperscript{2} Intel Labs \\
  \texttt{\small\{tam.nb.nguyen,dzung.doan,tat-jun.chin\}@adelaide.edu.au} \\
  \texttt{\small czptc2h@gmail.com}
}

\begin{document}
\maketitle
\begin{abstract}
Robust fitting of geometric models is a fundamental task in many computer vision pipelines. Numerous innovations have been produced on the topic, from improving the efficiency and accuracy of random sampling heuristics to generating novel theoretical insights that underpin new approaches with mathematical guarantees. However, one aspect of robust fitting that has received little attention is energy efficiency. This performance metric has become critical as high energy consumption is a growing concern for AI adoption. In this paper, we explore energy-efficient robust fitting via the neuromorphic computing paradigm. Specifically, we designed a novel spiking neural network for robust fitting on real neuromorphic hardware, the Intel Loihi~2. Enabling this are novel event-driven formulations of model estimation that allow robust fitting to be implemented in the unique architecture of Loihi~2, and algorithmic strategies to alleviate the current limited precision and instruction set of the hardware. Results show that our neuromorphic robust fitting consumes only a fraction (15\%) of the energy required to run the established robust fitting algorithm on a standard CPU to equivalent accuracy. 
\end{abstract}

\section{Introduction}

Many computer vision pipelines, ranging from visual SLAM, 3D reconstruction to image stitching, need to estimate geometric models from noisy and outlier-contaminated measurements that occur when operating in real-world environments~\cite{peter_meer_chapter}. Often, this is achieved by optimising the model parameters according to a robust criterion defined over the data, known as \textit{robust fitting}.

Random sampling heuristics such as RANSAC~\cite{fischler1981random} and its variants~\cite{s23010327} are established techniques for robust fitting. Such methods usually deliver satisfactory outcomes, but provide little insights on the veracity of the results~\cite{tran2014sampling}. On the other hand, techniques that rely on mathematical programming can provide some quality guarantees, but are typically too slow to be practical on real data~\cite{chin2018robust}.

Significant research has been devoted into robust fitting for computer vision. This includes improving the speed, accuracy and generalisability of random sampling methods~\cite{loransac,usac,graphcutransac} and deriving theoretical insights to better inform the design and usage of mathematical programming techniques (\eg, adapting to special cases, relaxing the guarantees)~\cite{pham2014interacting,le2019deterministic}. Machine learning approaches that can leverage statistics in training data to hypothesis sampling have also been developed~\cite{9008398}. Recently, quantum computing has also been explored for robust fitting~\cite{chin2020quantum}.

An aspect of robust fitting algorithm research that has received comparatively little attention is \emph{energy efficiency}. With the rapidly rising energy consumption of AI systems becoming a concern~\cite{energycrisis}, it is vital to develop vision algorithms that are energy-efficient. We argue that devising low-power alternatives for core components such as robust fitting is an essential step to enable ambitious low-power end-to-end 3D vision pipelines.

In this paper, we explore \textit{energy-efficient robust fitting} via neuromorphic computing, which is a bio-inspired computational model where a network of processing units called spiking neurons asynchronously send spike-based messages to each other~\cite{schuman2022opportunities}. Such a structure is called a spiking neural network (SNN). Due to the massive parallelism, stochastic behaviour, and event-driven computing, SNNs promise higher energy efficiency than conventional computing, including artificial neural networks (ANN)~\cite{Goodfellow-et-al-2016}.

Currently available neuromorphic processors that can implement SNNs include IBM TrueNorth~\cite{merolla2014million}, SpiNNaker~\cite{gonzalez2024spinnaker2} and Intel Loihi 1 and 2~\cite{davies2018loihi,orchard2021efficient, shrestha2024efficient}. Comprehensive experiments~\cite{davies2021advancing} indicate the much higher energy efficiency of the neuromorphic devices, which underlines their potential in reducing the energy consumption of data centres~\cite{intel_hala_point} and their application on embodied AI systems~\cite{mangalore2024neuromorphic}.

\vspace{-1em}
\paragraph{Contributions}

We develop an SNN that can conduct robust fitting on neuromorphic hardware, specifically Intel Loihi~2~\cite{shrestha2024efficient}. Underpinning our SNN are novel event-driven formulations of core steps in robust fitting, namely minimal subset sampling, model estimation and model verification, that make the problem amenable to a neuromorphic treatment. We also propose strategies to mitigate the current limitations on precision and instruction set of Loihi~2.

When simulated on the CPU, results of our SNN on synthetic data and real datasets verify its correctness and competitive accuracy relative to state-of-the-art robust fitting methods. Importantly, experiments on Loihi~2 illustrate the vastly superior energy efficiency of our SNN, in that it consumes only a fraction (15\%) of the energy required by established methods on the CPU.

\section{Related work}

\subsection{Computing paradigms for robust fitting}

Due to the crucial role of robust fitting in computer vision, various computing paradigms and hardware platforms have been explored to accelerate their execution.

FPGAs offer highly parallel computing capabilities, making them well-suited for real-time robust fitting tasks. Several works have explored robust fitting on FPGAs, leveraging the mechanism of parallel hypothesis evaluation to improve performance~\cite{vourvoulakis2016acceleration, vourvoulakis2017fpga, vourvoulakis2018fpga}. However, challenges such as limited on-chip memory, as well as difficulties in programming, optimising, and debugging algorithms, make FPGA implementations nontrivial~\cite{seng2021embedded}.

By devising differentiable versions of robust fitting and reformulating the problem as a machine learning task~\cite{brachmann2017dsac, brachmann2019neural, wei2023generalized}, GPUs have been adopted to conduct robust fitting. However, inferencing large neural networks on GPUs can be energy intensive~\cite{garcia2019estimation,georgiou2022green,alizadeh2024green}. Also, machine learning approaches suffer from generalizability issues if the testing data distribution differs from the training data distribution.

Recently, robust fitting using quantum computing has gained significant interest within the research community~\cite{chin2018robust, doan2022hybrid, yang2024robust}. These studies have shown promising potential for accelerating robust fitting processes. However, the quantum approaches remain in an experimental phase, with testing limited to small-scale data due to current hardware limitations. Furthermore, quantum computers have high energy demands due to cooling requirements~\cite{martin2022energy, national2018quantum}.

Unlike the above, we show that neuromorphic computing offers both high energy efficiency and usability.

\subsection{Neuromorphic computing for optimisation}

Research on algorithms in neuromorphic computing can be broadly categorized into two main directions: machine learning and optimization~\cite{schuman2022opportunities, davies2021advancing}. Learning approaches focus on converting pre-trained ANNs into corresponding SNNs for on-chip inference, directly learning SNN parameters through training equivalent proxy ANNs, and approximation of backpropagation on neuromorphic hardware. In the optimization domain, the spike-based temporal processing characteristics of SNNs are exploited to develop  solutions for optimization tasks. Our work belongs to the latter.

The temporal dynamics of SNNs have been actively explored in combinatorial and continuous optimization. In integer domains, a majority of handcrafted SNNs were designed for constraint satisfaction problems (CSPs), including travelling salesman problem, Sudoku, Boolean satisfiability and graph coloring~\cite{jonke2016solving, binas2016spiking, ostrau2019comparing, fonseca2017using, yakopcic2020leveraging}. The core idea behind these approaches is to encode CSP variables and constraints into an SNN topology. The SNN-based CSP solvers iteratively explore and refine the solution space, until seeking an assignment for variables that satisfy all constraints of the original CSP problem. Another line of focus in SNN-based combinatorial solvers is quadratic unconstrained binary optimization (QUBO)~\cite{alom2017quadratic, fang2022solving, mniszewski2019graph, corder2018solving, pierro2024solving}. While the primary objective of CSP solvers is to find feasible assignments to a combinatorial optimization problem, QUBO aims to find optimal solutions, where the inherent spike-based temporal dynamics of SNNs can be viewed as iterative solvers for optimization problems~\cite{theilman2025solving, mangalore2024neuromorphic, pierro2024solving}. 

\subsection{Neuromorphic computing for vision}

Neuromorphic computing is receiving increased attention in the vision community~\cite{schnider2023neuromorphic, chiavazza2023low, vitale2021event}.~\cite{schnider2023neuromorphic, chiavazza2023low} designed SNN-based for optical flow estimation from event stream data.~\cite{vitale2021event} integrated Loihi chip as a co-processor for computing angular error for drone control tasks. Neuromorphic computing has been explored in visual place recognition~\cite{Hussaini2025applications} and SLAM~\cite{Kreiser2018pose}. It is worthwhile to note that not all the works have tested on actual neuromorphic processors. Moreover, few have paid attention to robust fitting.

\section{Preliminaries}

We first review the problem formulation for robust fitting before giving a brief overview of SNNs and Intel Loihi 2.

\subsection{Robust fitting}\label{sec:classical_ransac}

We describe the procedure for robustly fitting a linear regression model, which is the specific problem targeted by our proposed SNN. This does not reduce the applicability of our ideas since many geometric models can be linearized~\cite[Chap.~4]{hartley2003multiple}. Moreover, our main aim is to establish the viability of neuromorphic robust fitting, and its extension to nonlinear models is left as future work.

Given a set of $N$ measurements $\mathcal{D} = \{(\mathbf{x}_i, y_i)\}_{i=1}^N$, where $\mathbf{x}_i \in \mathbb{R}^d$ and $y_i \in \mathbb{R}$, we wish to estimate the linear equation $y = \mathbf{x}^T \bm{\theta}$. The least squares (LS) solution is
\begin{align}\label{eq:ls}
    \hat{\btheta} = \argmin_{\bm{\theta} \in \mathbb{R}^d} \sum^{N}_{i=1} (y_i - \bx_i^T\btheta)^2 = \argmin_{\bm{\theta} \in \mathbb{R}^d} \| \bX\btheta -\by \|^2_2
\end{align}
where $\mathbf{X} \in \mathbb{R}^{N \times d}$ and $\mathbf{y} \in \mathbb{R}^N$ are obtained by vertically stacking $\{\mathbf{x}^T_i\}_{i=1}^N$ and $\{y_i\}_{i=1}^N$, respectively. If $\cD$ contains outliers, $\hat{\btheta}$ will be biased. Instead, robust fitting aims to find the model $\btheta$ that minimizes the objective function
\begin{equation}\label{eq:robust_formulation}
    \sum \limits_{i=1}^N \rho\left( \left| y_i - \bx^T_i \btheta  \right|\right),    
\end{equation}
where $\rho$ is a robust loss~\cite{zhang1997parameter}. Widely used in vision is
\begin{align}\label{eq:maxcon_rho}
    \rho(r) = \begin{cases} 1 & r > \epsilon_{\text{inlier}}, \\ 0 & \textrm{otherwise,} \end{cases}
\end{align}
where $\epsilon_{\text{inlier}}$ is the inlier threshold, hence, minimizing~\eqref{eq:robust_formulation} with~\eqref{eq:maxcon_rho} is equivalent to maximizing the inlier count
\begin{equation}\label{eq:max_con_formulation}
    \Psi(\bm{\theta}) = \sum \limits_{i=1}^N \mathbb{I}\left( \left| y_i - \bx^T_i \btheta  \right| \leq \epsilon_{\text{inlier}}\right)
\end{equation}
of $\btheta$, where $\mathbb{I}(\cdot)$ is an indicator function that returns 1 if the input condition is true and 0 otherwise. The value $\Psi(\bm{\theta})$ is called the consensus of $\btheta$~\cite{chin2017maximum}, and the resultant estimate is robust to outliers provided $\epsilon_{\text{inlier}}$ was selected appropriately.

Generating model hypotheses $\btheta$ by sampling minimal subsets is a successful approach for robust fitting~\cite{Mayo01051997}. Basically, three main steps are repetitively executed:
\begin{enumerate}
    \item Sample a $d$-subset $\mathcal{S} \subset \cD$ (minimal subset) of the data.
    \item Estimate a model hypothesis by LS fitting~\eqref{eq:ls} on $\cS$.
    \item Evaluate the quality of the hypothesis using~\eqref{eq:robust_formulation}.
\end{enumerate}
At termination, the algorithm returns the model with the best objective value. If the number of repetitions $K$ is large enough, at least one all-inlier minimal subset $\cS$ will be sampled, leading to a robust estimate of the model.

\subsection{SNN}\label{sec:SNN}

A neuromorphic algorithm can be designed as an SNN that is then executed on a neuromorphic computer~\cite{schuman2022opportunities}. Conceptually, an SNN consists of a set of $N$ spiking neurons, where each pair of neurons is connected via synapses. The synaptic connection strengths are represented by a weight matrix $\bW = [w_{ij}] \in \bbR^{N \times N}$, where $w_{ij} = 0$ indicates the absence of a connection between the $i$-th and $j$-th neurons.

Each spiking neuron comprises internal states that accumulate input stimuli over time, and emits spikes only when predefined conditions are met. Upon spiking, it transmits a spike signal to connected neurons and may enter into a refractory (inactive) period. Well-known spiking neuron models include Leaky Integrate-and-Fire~\cite{gerstner2002spiking}, Resonate-and-Fire~\cite{izhikevich2001resonate} and Izhikevich~\cite{izhikevich2003simple}, though the SNN framework is flexible enough to allow custom neurons with specific computations. The synaptic weights and neuronal processing define the problem that is solved by the SNN.

\subsection{Intel Loihi 2}\label{sec:hardwareprelim}

\begin{figure}[h]
    \centering
    \includegraphics[width=0.99\columnwidth]{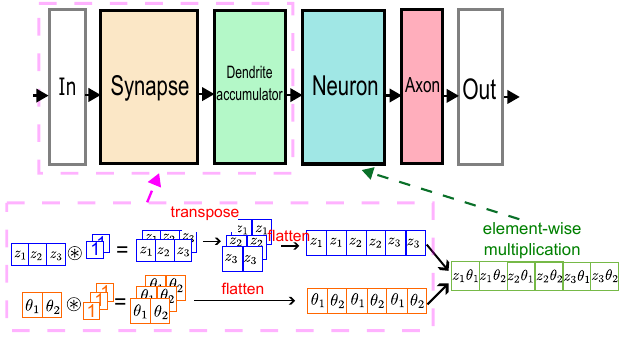}
    \caption{The top diagram shows a high-level schematic of a neuro core in Loihi 2. Note that Neuron block is programmable. The bottom diagram shows our technique to emulate matrix-matrix multiplication and how it is mapped onto the neuro core. See Fig.~AA Supplementary for more details.}
    \label{fig:neurocore}
\end{figure}

Each Loihi 2 chip~\cite{orchard2021efficient} houses 128 asynchronous neuromorphic cores (neuro cores) which can simulate up to 8192 stateful and parallel spiking neurons. At each neuro core, the ingress spikes from other cores enter a \textit{Synapse} block, where a dense/ sparse matrix-vector multiplication or convolution is performed; see Fig.~\ref{fig:neurocore}. The result is then accumulated and passed as inputs to the \emph{Neuron} block. Here, stateful neuron programs are executed and 24-bit spike messages are generated and routed to other neuro cores~\cite{davies2018loihi, mangalore2024neuromorphic}. The SNN architecture can be designed using \textit{Dense}, \textit{Sparse}, and \textit{Convolution} synaptic configurations of 8-bit weights, and the neuron models can be customised through assembly code with up to 24-bit internal states.

\section{Neuromorphic robust fitting}\label{sec:neuromorphic_ransac}

Here, we describe the proposed SNN for neuromorphic robust fitting, including the mathematical underpinnings, spiking neuron designs, and hardware implementation details. Note that while spiking neurons are conceptually asynchronous or event-driven~\cite{schuman2022opportunities}, the operations of neurons below are described using discrete timesteps $t$ to give more intuitive depiction of time evolution. This matches the programming at \emph{the neuron level} in real neuromorphic computers such as Loihi~2, which are fully digital devices~\cite{davies2018loihi}. However, it should be reminded that $t$ is the \emph{algorithmic time} rather than a global synchronous clock.

\subsection{SNN for robust fitting}\label{sec:snn_for_ransac}

\begin{figure}[ht]\centering
    \includegraphics[width=0.99\columnwidth]{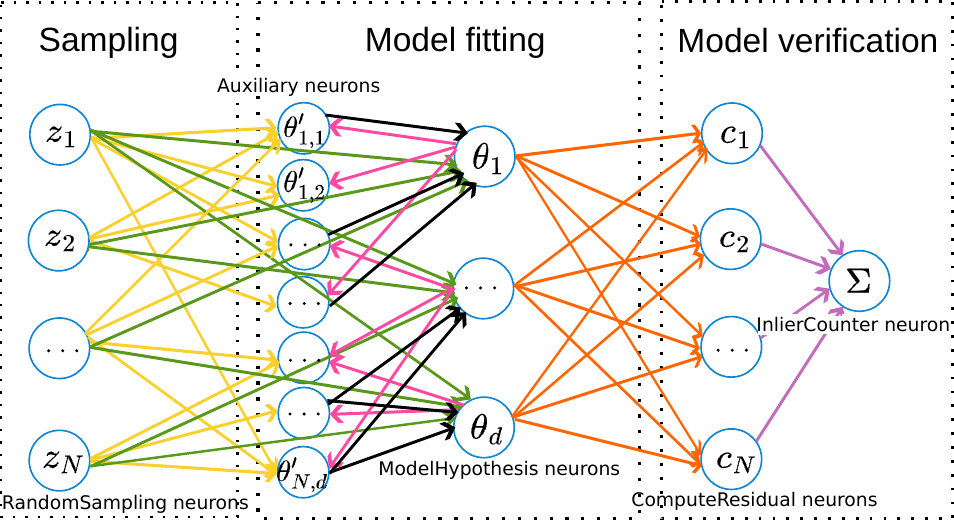}  
    \caption{The proposed NeuroRF SNN.}
    \label{fig:ransacsnn}
\end{figure}

Fig.~\ref{fig:ransacsnn} illustrates the proposed SNN for robust fitting, called \emph{NeuroRF}, which consists of the following neurons:
\begin{itemize}
    \item $N$ RandomSampling neurons $\{ z_i \}^{N}_{i=1}$.
    \item $d$ ModelHypothesis neurons $\{ \theta_j \}^{d}_{j=1}$.
    \item $Nd$ Auxiliary neurons $\{ \theta^\prime_{i,j} \}^{j=1,\dots,d}_{i=1,\dots,N}$.
    \item $N$ ComputeResidual neurons $\{ c_j \}^{N}_{j=1}$.
    \item $1$ InlierCounter neuron $\Sigma$.
\end{itemize}

Algorithm~\ref{algo:SNN5_revised_algorithm_cont} defines the internal arithmetic operations that evolve the state of each neuron. Note that, unlike in the case of classical (von Neumann) computers, there is no clear main body and parent-subfunction relationships, though the neurons influence each other by sending output spikes through the interconnections.

The behaviour of NeuroRF will be described in more detail below via its main operations, while details of implementation on Loihi 2 will be provided in Sec.~\ref{sec:hardware}.

\subsubsection{Random sampling} 

As the name suggests, the role of the RandomSampling neurons is conducting random sampling of the input data. Each $z_i$ is a binary variable, where $z_i = 1$ means that the $i$-th point $(\bx_i,y_i)$ is selected and $z_i = 0$ means otherwise. 

For a $d$-dimensional model, each neuron emits a spike with probability $d/N$. This is encoded in a simple dynamic by comparing a randomly generated number with a constant $d/N$. However, since the neurons sample independently, they may not collectively select exactly $d$ points. Since we conduct estimation using gradient descent (details to follow), over- or under-sampling the points (including selecting no points) do not cause numerical issues.

\subsubsection{Model hypothesis generation}

The LS objective~\eqref{eq:ls} can be written in the quadratic form
\begin{equation}
    f(\bm{\theta}) = \bm{\theta}^T \bf{Q} \bm{\theta} + \bf{p}^T \bm{\theta},
    \label{eq:ls_as_qp}
\end{equation}
where $\bQ = \bX^T\bX$ and $\bp^T = -\by^T \bX$. Since $f(\btheta)$ is convex, it justifies using gradient descent (GD) to solve least squares, where the first-order gradient is
\begin{equation}\label{eq:nabla_f_theta}
    \nabla f (\bm{\theta}) = \bQ \bm{\theta} + \bp,
\end{equation}
and we iteratively update $\btheta$ via
\begin{equation}\label{eq:gd}
    \bm\theta^{(t)} = \bm\theta^{(t-1)} - \alpha \left(\bQ \bm\theta^{(t-1)} + \bp\right)   
\end{equation}
with $\alpha$ being the step size or learning rate for $M$ iterations.

Previous studies have shown that the spike-based temporal dynamics of SNNs align with the behaviour of classical iterative solvers~\cite{theilman2025solving, mangalore2024neuromorphic, pierro2024solving, nguyen2025slack}. In fact, the GD iteration~\eqref{eq:gd} readily lends itself as the dynamic equations for the ModelHypothesis neurons, where each evolves according to

\begin{equation}
    \theta_j^{(t)} = \theta_j^{(t-1)} - \alpha \left( \bq_{j} \btheta^{(t-1)} + p_j\right)
    \label{eq:gd_neuron},
\end{equation}
with $\bq_{j}$ being the $j$-th row of $\bQ$ and $p_j$ being the $j$-th element of $\bp$. Conceptually, $\bq_{j}$ and $p_j$ are respectively the synaptic weights and bias leading into the $j$-th neuron.

However, a direct application of~\eqref{eq:gd_neuron} is problematic for our aims, since the data for estimation is encoded in $\bQ$ and $\bp$, which need to be repetitively sampled at each robust fitting iteration, while the synaptic weights and biases are constant in the Synapse block of Loihi 2 during operation.

We propose an algebraic manipulation that enables fully neuromorphic random sampling and model hypothesis generation.

Collecting the RandomSampling states in a vector $\bz = \left[z_1, z_2, \dots, z_N \right]^T$, we ``lift'' the gradient~\eqref{eq:nabla_f_theta} as
\begin{equation}
    \nabla f (\bm{\theta}, \bz) = \bQ^{\prime} \bm{\theta}^{\prime} + \bP^{\prime} \bz
    \label{eq:nabla_f_theta_z_method}
\end{equation}
where
\begin{align}\label{eq:Q_prime_method}
\begin{aligned}
    \bQ^{\prime} &= \left[ \bx_1 \bx_1^T \quad \bx_2  \bx_2^T \quad \ldots \quad \bx_N \bx_N^T \right] \in \bbR^{d \times Nd},\\
    \bm{\theta}^{\prime}  &= \operatorname{vec} \left( \bz \bm{\theta}^T \right) = \bz \otimes \bm{\theta} \in \bbR^{Nd},\\
    \bP^{\prime} &= \left[ -y_1 \mathbf{x_1} \quad -y_2 \mathbf{x_2} \quad \ldots \quad -y_N \mathbf{x_N} \right] \in \mathbb{R}^{d \times N}.    
\end{aligned}
\end{align}
Note that the $\otimes$ is the Kronecker product. The lifting allows the gradient to be a function of the data selection via $\bz$. The GD update is now also dependent on $\bz$, \ie, 
\begin{equation}
    \bm \theta^{(t)} = \bm \theta^{(t - 1)} - \alpha\left( \bQ^{\prime} \bm\theta^{\prime (t-1)} + \bP^{\prime} \bz \right).
    \label{eq:gd_with_z}
\end{equation}
See Sec.~A Supplementary for details of the derivation. Note that the extreme cases ($\bz = \mathbf{1}$ and $\bz = \mathbf{0}$) do not cause numerical issues, since the gradient reduces to $\nabla f(\btheta)$ and $\mathbf{0}$ respectively (no operation in the latter case).

The expansion~\eqref{eq:Q_prime_method} creates $Nd$ Auxiliary neurons $\btheta^\prime$, which are unrolled into $\{ \theta^\prime_{i,j} \}^{j=1,\dots,d}_{i=1,\dots,N}$. Each Auxiliary neuron evolves according to
\begin{align}
    \theta^{\prime (t)}_{i,j} = z_i^{(t-1)} \theta_j^{(t-1)},
    \label{eq:theta_prime_dynamics}
\end{align}
which behaves as a coupling of a pair of RandomSampling and ModelHypothesis neurons. More importantly, the ModelHypothesis neurons now follow the dynamical equation
\begin{equation}\label{eq:gd_neuron_Q_prime_P_prime}
    \theta_j^{(t)} = \theta_j^{(t-1)} - \alpha \left( \bq^\prime_{j} \btheta^{\prime (t-1)} + \bp'_{j}\bz\right),
\end{equation}
where $\bq^\prime_{j}$ is the $j$-th row of $\bQ^\prime$ and $\bp^\prime_j$ is the $j$-th row of $\bP^\prime$. The synaptics weights and biases leading into the $j$-th ModelHypothesis Neuron $\theta_j$ are now \emph{constant}.

\subsubsection{Model verification}

The ComputeResidual and InlierCounter neurons calculate the residuals and consensus~\eqref{eq:max_con_formulation} of the current model estimate (states of the ModelHypothesis neurons). These are straightforward, and the reader is referred to Algorithm~\ref{algo:SNN5_revised_algorithm}.

\begin{algorithm}[t]
    \caption{NeuroRF algorithm}
    \label{algo:SNN5_revised_algorithm}
    \begin{algorithmic}[1]
        \BlockTitle{RandomSampling layer}
        \Require Switching probability $prob = \frac{d}{N}$
        \Require $\tau = 2M + 4, \quad \mathcal{L} = K * \tau$
        \State Initialize $k \gets 0, \quad counter \gets 0$
        \State Initialize $\mathbf{z}^{(0)} \gets \mathbf{0}_{\text{N}}$
        \vspace{1mm}
        \For{$t = 1, 2,\ldots,\mathcal{L}$}
            \State $counter \mathrel{+}= 1$
            \State $next\_sampling\_idx = \tau*k + 1$

            \vspace{1mm}
            \If{$counter = next\_sampling\_idx$}
                \State $k \mathrel{+}= 1$
                \State $\bm{\gamma}^{(t)} = \operatorname{rand}[0, 1)$
                \State $\bz^{(t)} = prob \geq \bm{\gamma}^{(t)}$
                
            \Else
                \State $\bz^{(t)} = \bz^{(t-1)}$
            \EndIf
            \State Send $\bz^{(t)}$ to connected Auxiliary and ModelHypothesis layers.
        \EndFor
    \end{algorithmic}
    
    \vspace{2mm}

    \begin{algorithmic}[1]
        \BlockTitle{Auxiliary layer}
        \Require Connection matrices $\bF_{d}$, $\bF_{N}$
        
        \Require $\tau = 2M + 4, \quad \mathcal{L} = K * \tau$
        \State Initialize $\bm{\theta}^{\prime(0)} \gets \bm{0}_{\text{Nd}}$
        \State Initialize $k \gets 1, \quad counter \gets 0$

        \vspace{1mm}
        \For{$t = 1, 2,\ldots,\mathcal{L}$}

            \State $\bz\_{\text{in}}^{(t)} = \bF_{d} \circledast \bz^{(t-1)}$
            \State $\bm{\theta}\_{\text{in}}^{(t)} = \bF_{N} \circledast \bm{\theta}^{(t-1)}$
            
            \State $counter \mathrel{+}= 1$
            \State $next\_sampling\_idx = \tau*k + 1$
            \vspace{1mm}

            \If{$counter = next\_sampling\_idx$}
                \State $k \mathrel{+}= 1$
                \State $\bm{\theta}^{\prime(t)} = \bm{0}_{\text{Nd}}$
            \Else
                \State $\bm{\theta}^{\prime(t)} = \mathbf{z}\_\text{in}^{(t)} * \bm{\theta}\_\text{in}^{(t)}$

            \EndIf
            \State Send $\bm{\theta}^{\prime(t)}$ to connected ModelHypothesis layer.
        \EndFor
    \end{algorithmic}

\end{algorithm}

\addtocounter{algorithm}{-1}
\begin{algorithm}[t]
    \caption{NeuroRF algorithm (cont.) }
    \label{algo:SNN5_revised_algorithm_cont}
    \begin{algorithmic}[1]
        \BlockTitle{ModelHypothesis layer}
        \Require Connection matrices $\mathbf{Q}^{\prime}$, $\mathbf{P}^{\prime}$ and learning rate $\alpha$ 
        \Require $\tau= 2M + 4, \quad \mathcal{L} = K * \tau$
        \vspace{1mm}
        
        \State Initialize $\bm{\theta}^{(0)} \gets \bm{0}_{\text{d}}$
        \State Initialize $k \gets 1, \quad counter \gets 0$
        
        \vspace{1mm}
        \For{$t = 1, 2,\ldots,\mathcal{L}$}

            \State a\_in$^{(t)} = \mathbf{Q}^{\prime} \bm{\theta}^{\prime(t-1)}$
            \State bias$^{(t)} = \mathbf{P}^{\prime} \mathbf{z}^{(t-1)}$
            
            \State $counter \mathrel{+}= 1$
            \State $next\_sampling\_idx = \tau*k + 1$

            \If{$counter = new\_sampling\_idx$}
                \State $k \mathrel{+}= 1$
                \State $\bm{\theta}^{(t)} = \mathbf{0}_{\text{d}}$
            \Else
                \If{$counter$ is odd}
                    \State total\_grad $^{(t)}$ = a\_in$^{(t)}$ + bias$^{(t)}$

                    \State $\bm{\theta}^{(t)} = \bm{\theta}^{(t-1)} - \alpha * \text{total\_grad}^{(t)}$

                \Else
                    \State $\bm{\theta}^{(t)} = \bm{\theta}^{(t-1)}$
                \EndIf
            \EndIf
            \State Send $\bm{\theta}^{(t)}$ to Auxiliary layer
        \EndFor
    \end{algorithmic}

    \vspace{2mm}

    \begin{algorithmic}[1]
        \BlockTitle{ComputeResidual layer}
        \Require Connection matrix $\mathbf{X}$, bias vector $\mathbf{y}$ and inlier threshold $\epsilon_{\text{inlier}}$
        \Require $\tau= 2M + 4, \quad \mathcal{L} = K * \tau$

        \vspace{1mm}
        
        \State{Initialize $\mathbf{c} \gets \mathbf{0}_{\textbf{N}}$}
        \vspace{1mm}

        \For{$t = 1,2,\ldots, \mathcal{L}$}
            \State $\text{a\_in}^{(t)} = \mathbf{X} \bm{\theta}^{(t-1)}$
            \State $\mathbf{res}^{(t)} = \text{a\_in}^{(t)} - \mathbf{y}$
            \State $\mathbf{c}^{(t)} = |\mathbf{res}^{(t)}| \leq \epsilon_{\text{inlier}}$
            
            \State Send $\bc^{(t)}$ to InlierCounter neuron

        \EndFor
    \end{algorithmic}

    \begin{algorithmic}[1]
        \BlockTitle{InlierCounter neuron}
        \Require Connection matrix $\bm{1}^T_{\text{N}}$
        \Require $\tau= 2M + 4, \quad \mathcal{L} = K * \tau$

        \vspace{1mm}
        \For{$t = 1,2,\ldots, \mathcal{L}$}
            \State $\Psi^{(t)} = \mathbf{1}^T_N \mathbf{c}^{(t-1)}$
        \EndFor

    \end{algorithmic}
\end{algorithm}

\subsection{Mitigating hardware limitations}\label{sec:hardware}

To implement NeuroRF on Loihi 2~\cite{shrestha2024efficient}, the SNN architecture and synaptic weights are defined on a host CPU, then mapped to the neuro cores of Loihi 2 via the Lava framework~\cite{Lava}. Consensus sets found are read out and returned to the host CPU once the algorithm finishes. This probe is performed by special embedded processors that can access the states of neuro cores~\cite{evanusa2019event, chiavazza2023low}. 

However, Loihi 2 has several constraints: 8-bit weights which can limit the instance size for testing, no support for floating-point arithmetic, a reduced instruction set (no division) and synapse configurations with limited customisability~\cite{intel_brief_instruction_set, pierro2024solving, mangalore2024neuromorphic}. Here, we  present strategies to mitigate these constraints. Note that the limitations may be solved in future iterations of Loihi, and do not detract from the mathematical and logical consistencies of NeuroRF.

\vspace{-1em}
\paragraph{Integer arithmetic}

All data $\cD$ is assumed integer or converted to integer from rational numbers. The optimal model estimate $\btheta$ may not be integral, but NeuroRF is aimed at finding the best integer solution. As we will show in Sec.~\ref{sec:results}, the integer approximation does not significantly reduce the quality (consensus size) obtained, as observed in other applications on Loihi 2~\cite{theilman2025solving, mangalore2024neuromorphic}.

\vspace{-1em}
\paragraph{Random sampling} Loihi 2's pseudo-random generator (PRG) is restricted to either 16 or 24 bit integers~\cite{pierro2024solving}. We describe our approach to using 16-bit PRG, but the idea is similar for the 24-bit PRG. Let $LFSR \in [0, 2^{16} - 1]$ be the Loihi 2-generated random number. The required threshold $\gamma \in [0, 1)$ is mathematically obtained as
\begin{equation}
    \gamma = \frac{LFSR}{2^{16} - 1} \approx \frac{LFSR}{2^{16}}.
\end{equation}
From Algorithm~\ref{algo:SNN5_revised_algorithm}, the switching condition for a RandomSampling neuron is mathematically defined as
\begin{equation}
    d > \frac{N * LFSR}{2^{16}}.
    \label{eq:convert_switching_condition}
\end{equation}
We approximate the division using the arithmetic right shift, leading to the switching logic
\begin{equation}
    d > ((N * LFSR) \gg 16).
    \label{eq:loihi_switching_condition}
\end{equation}

\vspace{-1em}
\paragraph{GD update step size} The GD process~\eqref{eq:gd_with_z} requires a step size parameter $\alpha$, which should be sufficiently small to not cause divergence. To allow a floating-point $\alpha$, we use fixed-point representation~\cite{mangalore2024neuromorphic,theilman2025solving} for storing $\alpha$. Specifically, an $\alpha \in \bbR$ can be converted to a fixed-point counterpart
\begin{equation}
    \bar\alpha = \operatorname{ceil}(\alpha * 2^{\beta}),
    \label{eq:alpha_floating_to_fixed_pt}
\end{equation}
where $\beta \in \mathbb{Z}$ is a tuned constant. Here $\bar\alpha$ is rounded up to the nearest integer. The update~\eqref{eq:gd_with_z} can be approximated as
\begin{equation}
    \bm \theta^{(t)} = \bm \theta^{(t - 1)} - \left( \bQ^{\prime} \bm\theta^{\prime (t-1)} + \bP^{\prime} \bz \right) \frac{\bar \alpha}{2^{\beta}}.
    \label{eq:gd_with_z_alpha_bar}
\end{equation}
Using the arithmetic right shift, the GD update becomes
\begin{equation}
    \bm \theta^{(t)} = \bm \theta^{(t - 1)} - \left[ \left( \bQ^{\prime} \bm\theta^{\prime (t-1)} + \bP^{\prime} \bz \right) * \bar \alpha \right] \gg  \beta
    \label{eq:gd_with_z_loihi}
\end{equation}
This approximation at each update step could cause overall drift of the solution~\cite{theilman2025solving, mangalore2024neuromorphic}, thus the number of steps should be kept small to avoid large build-up of errors.

\vspace{-1em}
\paragraph{Matrix-matrix multiplications}

On Loihi 2, neurons must connect through Synapse (See Fig~\ref{fig:neurocore}), where an interaction $\bm{\theta}^{\prime} = \operatorname{vec}(\bz\bm{\theta}^T)$ is not feasible. To realise the Auxiliary neural dynamics, we introduce two synaptic connection matrices into $\bm\theta^{\prime}$
\begin{equation}
    \bm{\theta}^{\prime} = \operatorname{vec} \left[ \left( \mathbf{1}_{\text{d}} \mathbf{z}^T \right)^T \right] * \operatorname{vec} \left( \mathbf{1}_{\text{N}} \bm{\theta}^T\right).
    \label{eq:derive_theta_prime}
\end{equation}

Here $\mathbf{1}_{\text{d}}$ and $\mathbf{1}_{\text{N}}$ can be viewed as synaptic weights of matrix-matrix multiplication Synapses, which are not supported on Loihi 2. To overcome this limitation, we emulate matrix-matrix multiplication with \textit{Conv} Synapse
\begin{equation}
    \left( \mathbf{1}_{\text{d}} \bz^T \right) = \bF_{\text{d}} \circledast \bz, \quad \left( \mathbf{1}_{\text{N}} \bm{\theta}^T\right) = \bF_{\text{N}} \circledast \bm{\theta},
\end{equation}
where $\circledast$ is a convolution operation. $\bF_{N} \in \bbR^{N \times 1 \times 1}$ and $\bF_{d} \in \bbR^{d \times 1 \times 1}$ are respectively the set of $N$ and $d$ $(1 \times 1)$ convolutional filters. See Fig.~\ref{fig:neurocore} for a visual explanation of our approach to emulate matrix-matrix multiplication and how this process is mapped onto neuro cores.

\section{Results}\label{sec:results}

Our experiments focused on establishing the correctness of NeuroRF, its performance on a neuromorphic chip, and its potential in lowering the energy cost of robust fitting. 

\subsection{Method variants}

Two different variants of NeuroRF were used:
\begin{itemize}
    \item \textbf{NeuroRF-CPU}: NeuroRF was implemented in Lava~\cite{Lava} with 64-bit floating-point arithmetic and x86-64 instruction set, and run on an Intel Core i7-11700K. This simulated NeuroRF on a CPU.
    \item \textbf{NeuroRF-Loihi}: NeuroRF was implemented on Loihi 2 (specifically on the single-chip Oheo Gulch board~\cite{intel_brief_instruction_set}) using Lava-Loihi 0.7.0~\cite{Lava} and Loihi assembly language~\cite{orchard2021efficient}. Note that Loihi 2 had limited precision and instruction set, which we mitigated in Sec.~\ref{sec:hardware}.

\end{itemize}

\subsection{Correctness of NeuroRF}
\label{sec:correctness_neurors}

We first validate NeuroRF on robust linear regression via synthetic data. For each data instance, we randomly generated a true model $\bm{\theta}^*$ and $N$ independent measurements $\{\bx_i\}_{i=1}^N$. Each $y_i$ was then computed based on linear model $\by = \bm{\theta}^T \bx$ and perturbed with Gaussian noise of $\sigma_{\text{inlier}} = 0.1$. To simulate outliers, $\eta\%$ data points were randomly corrupted with Gaussian noise of $\sigma_{\text{outlier}} = 1.5$.  The inlier threshold $\epsilon_{\text{inlier}}$ was set to $0.5$.

To rigorously test our method, we generated 65 problem instances of varying levels of difficulty, with 5 instances produced for each combination of $(N, d, \eta)$:
\begin{itemize}
    \item $N \in [100, 200, 400, 300,500]$ with $d = 8$ and outlier ratio $\eta = 20$.
    \item $d \in [2, 3, 6, 8]$ with $N = 200$ and $\eta = 20$.
    \item $\eta \in [10, 20, 30, 40, 50, 60]$ with $N = 200$ and $d = 8$.
\end{itemize}

We compared NeuroRF and a classical robust fitting algorithm RANSAC~\cite{fischler1981random}, denoted as \textbf{RS-CPU}, running on an Intel Core i7-11700K. The number of iterations $K$ was set to $300$ for both methods, while $M = 200$ and $\alpha = 0.02$ were configured for \neurorscpu. Since the goal here is to verify the correctness of our SNN on synthetic data, it is sufficient to compare against RS-CPU. We will benchmark against more advanced methods on real data in Sec.~\ref{sec:affine_results}.

To evaluate robust fitting accuracy, we employed the normalized Euclidean distance between the ground-truth and output models, $\btheta_{gt}$ and $\btheta_{est}$, from a robust fitting method:
\begin{equation}\label{eq:normdist}
    100 * \lVert \btheta_{gt} - \btheta_{est} \rVert_2/\lVert \btheta_{gt}\rVert_2.
\end{equation}
We recorded the average and std.~dev.~ of the normalized distance over 10 trials for each method.

\begin{figure*}[t]\centering
\begin{subfigure}[b]{0.33\textwidth}
    \includegraphics[width=0.99\columnwidth]{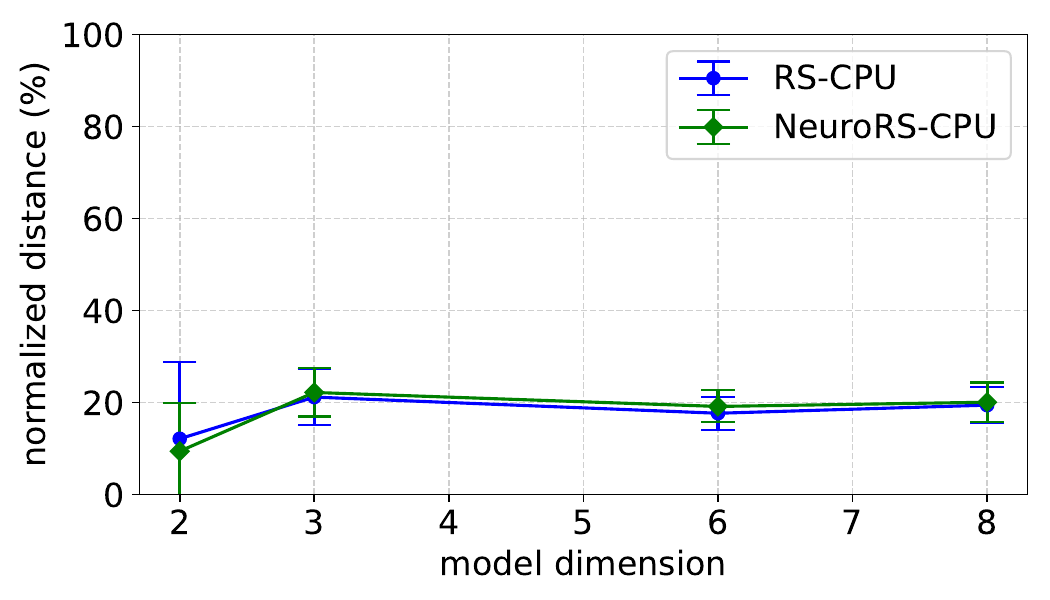}
    \caption{Effect of dimension $d$  $(N=200, \eta = 0.2)$}
    \label{fig:rf_wrt_d}
\end{subfigure}
\begin{subfigure}[b]{0.33\textwidth}
    \includegraphics[width=0.99\columnwidth]{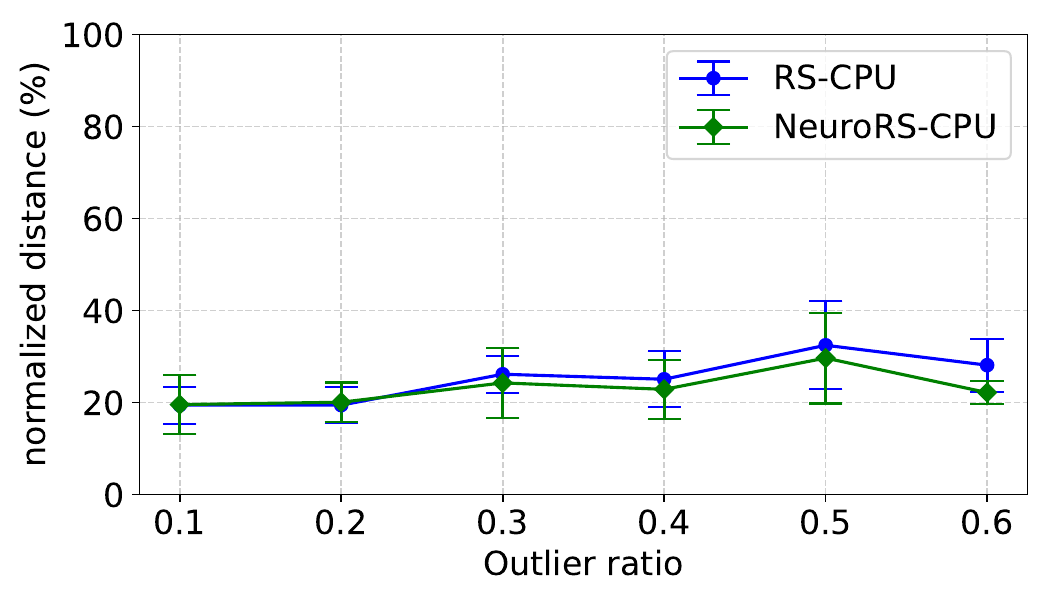}
    \caption{Effect of outlier ratio $\eta$ $(N=200, d = 8)$ }
    \label{fig:rf_wrt_eta}
\end{subfigure}
\begin{subfigure}[b]{0.33\textwidth}
    \includegraphics[width=0.99\columnwidth]{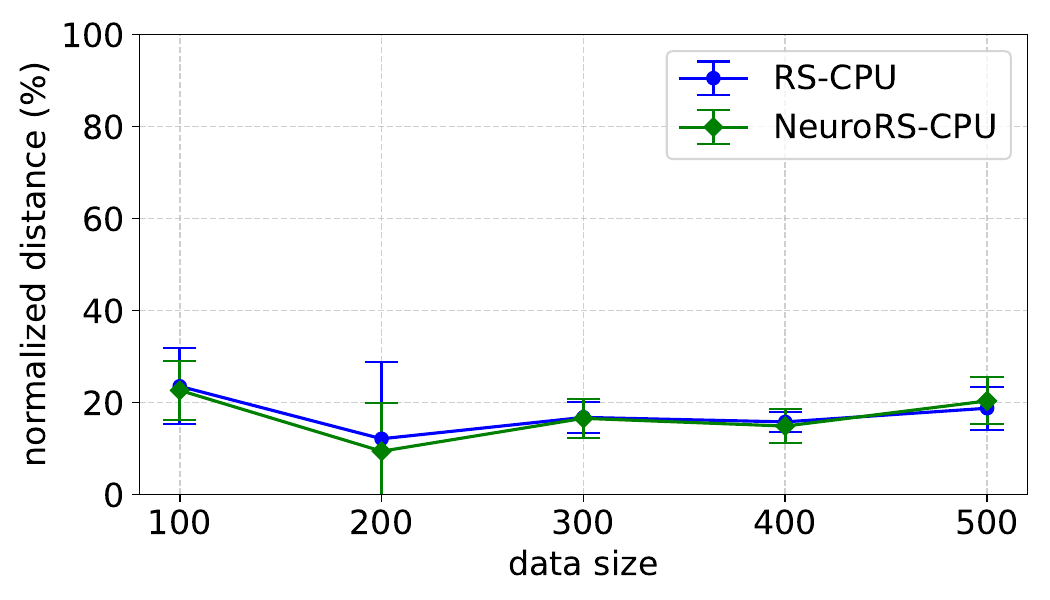}
    \caption{Effect of data size $N$ $(\eta = 0.2, d = 8)$}
    \label{fig:rf_wrt_N}
\end{subfigure}
\vspace{-1em}
\caption{Normalized Euclidean distance $(\%)$ across various levels of difficulty. Results were averaged over 10 trials for each method.}
\label{fig:rf_on_cpu}
\end{figure*}

Fig.~\ref{fig:rf_on_cpu} plots the normalized distance across the $N$, $d$ and $\eta$ variations. \neurorscpu~achieved comparable performance to \rscpu~on the three setups, which confirms the algorithmic soundness of NeuroRF and its effectiveness in handling high-dimensional problems. Also, see Sec.~C Supp. for runtime results. Note that the runtime figures here were from \emph{simulating} NeuroRF on a CPU, and hence are not reflective of runtime on Loihi 2.

\subsection{Performance on neuromorphic hardware}

Due to the relatively low capacity of Loihi 2, we tested~\neurorsloihi~on line fitting problems only ($d = 2$). 

Since the synaptic weights are 8 bits only, we generated synthetic data as follows: in each instance, the true model $\btheta^\ast$ was sampled from $\{-10,-9,\dots,9,10\}^2$, from which $N \in \{10, 20\}$ integer points that satisfy the linear relation were generated. All points were randomly corrupted with noise of $\pm 1$, before $\eta \in \{10,20,\ldots, 50\}$ percent were selected and corrupted further with noise of $\pm 4$. For each $(N,\eta)$ combination, we generated 5 instances, leading to a total of 50 line fitting problems.

We compared~\neurorsloihi~and~\rscpu; the more advanced methods (Sec.~\ref{sec:affine_results}) were not more energy-efficient than \rscpu. For both methods, $\epsilon_{\text{inlier}}$ was set to $4$ and the number of iterations $K$ was set to $100$. For~\neurorsloihi, we also set $M = 200$, $\alpha=0.02$ and $\beta = 10$.

\begin{figure*}[t]\centering
\begin{subfigure}[b]{0.33\textwidth}
    \includegraphics[width=0.99\columnwidth]{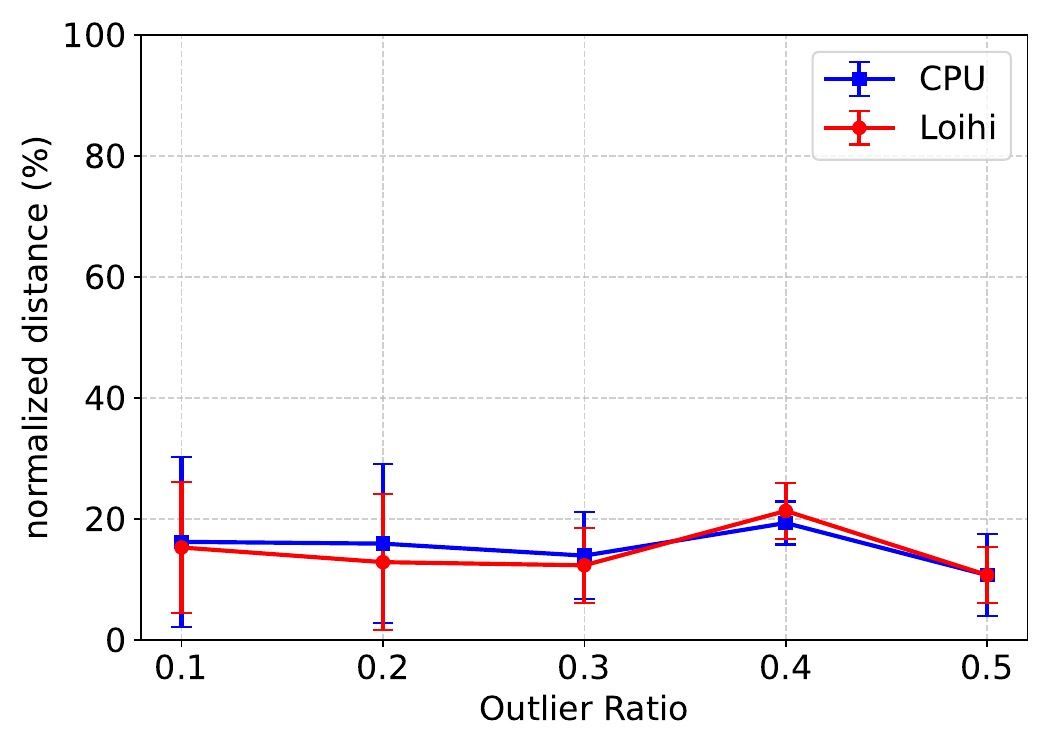}
    \caption{}
    \label{fig:consensus_size_N_20}
\end{subfigure}
\begin{subfigure}[b]{0.33\textwidth}
    \includegraphics[width=0.99\columnwidth]{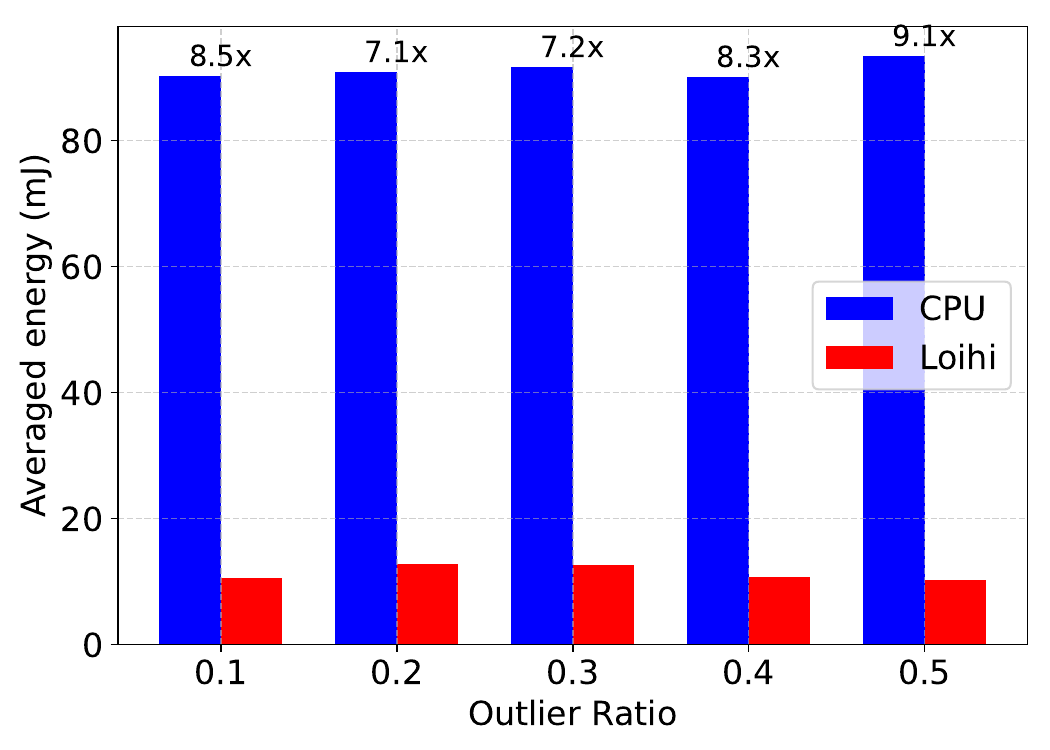}
    \caption{}
    \label{fig:energy_consumption_N_20}
\end{subfigure}
\begin{subfigure}[b]{0.33\textwidth}
    \includegraphics[width=0.99\columnwidth]{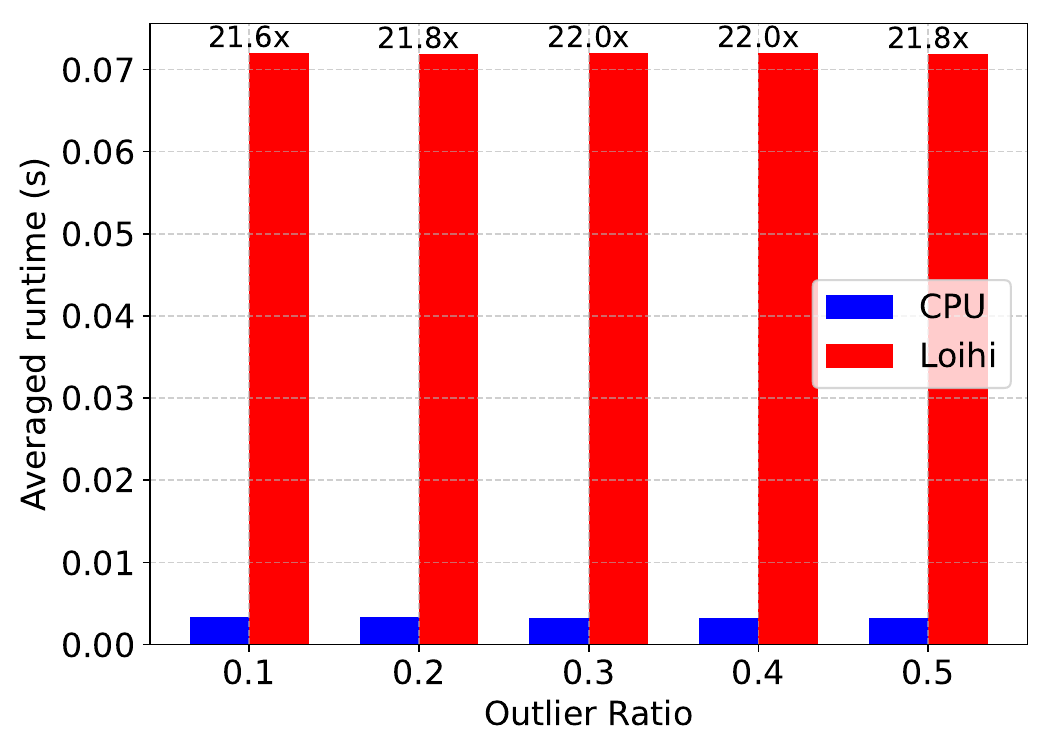}
    \caption{}
    \label{fig:runtime_N_20}
\end{subfigure}

\vspace{-1em}
\caption{Performance on neuromorphic hardware. (a) Normalized Euclidean distance $(\%)$, (b) average dynamic energy consumption and (c) average runtime of \neurorsloihi~and RS-CPU on synthetic line fitting instances with $N = 20$ points, plotted against outlier rate.}
\label{fig:hardwareperformance}
\end{figure*}

We recorded the distance~\eqref{eq:normdist}, runtime and dynamic energy consumption of the methods. Results were averaged over 10 trials and the std.~dev.~was also provided in the case of consensus size. For RS-CPU, runtime was obtained with \textit{time} python module while the energy consumption was measured with Intel's RAPL technology via pyJoules~\cite{pyjoules}. For~\neurorsloihi, the energy consumption and runtime were obtained via the built-in Loihi 2 Profiler.

Fig.~\ref{fig:hardwareperformance} shows results for $N = 20$ (plots for $N = 10$ are available in Sec.~D Supp). Fig.~\ref{fig:consensus_size_N_20} shows the normalized distance in percentage of~\neurorsloihi~and~\rscpu, which clearly indicates that both are on par in terms of soluton quality. Fig.~\ref{fig:energy_consumption_N_20} shows that~\neurorsloihi~consumed about only $15 \%$ of the energy required by~\rscpu, a definitive proof of the much higher energy-efficiency of~\neurorsloihi. However, Fig.~\ref{fig:runtime_N_20} shows that the runtime of~\neurorsloihi~was greater than~\rscpu. This was probably because~\neurorsloihi~employed GD for LS, while~\rscpu~solved LS analytically. Also, the speed of each Loihi~2 node is not as high as a cutting-edge CPU.

\subsection{Affine image registration}\label{sec:affine_results}

We illustrate the applicability of NeuroRF to robust fitting on real visual data through an image registration problem.

An affine transformation $\bH_{\bA}$ warps a point $\bx$ in one image to a corresponding point $\bx^\prime$ in another image via
\begin{equation}
    \mathbf{x}^{\prime} = \mathbf{H}_\text{A} \widetilde{\mathbf{x}},
    \label{eq:affine_transform}
\end{equation}
where $\mathbf{x}^{\prime} = (x', y')$ and $\widetilde{\mathbf{x}} = [\mathbf{x}^T \quad 1] = (x, y, 1)$ is $\mathbf{x}$ in homogeneous coordinates. The $2 \times 3$ affinity matrix $\mathbf{H}_\text{A}$ can be estimated by solving a linear system constructed from 3 point correspondences~\cite[Chap.~2]{hartley2003multiple}. Each correspondence $\langle \mathbf{x}'_i, \mathbf{x}_i \rangle$ contributes two equations, forming a system of six equations to estimate 6 parameters. The model dimension $d$ of the affine image registration problem is hence 6. When outliers are present in the correspondences, robust estimation methods can be applied with a minimal subset size of 3. For the conversion to~\eqref{eq:max_con_formulation} and how to apply our method to affine transformation, see Sec.~B Supp.

\begin{figure}[h]
    \centering
    \begin{subfigure}[t]{\linewidth}
        \centering
        \includegraphics[width=0.99\linewidth,height=9em]{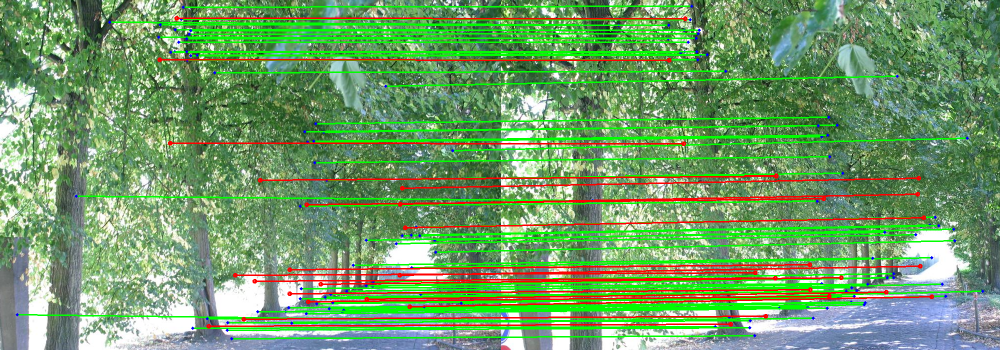}
        \caption{Trees}
        \label{fig:first_image}
    \end{subfigure}

    \vspace{0.1em} 

    \begin{subfigure}[t]{\linewidth}
        \centering
        \includegraphics[width=0.99\linewidth,height=9em]{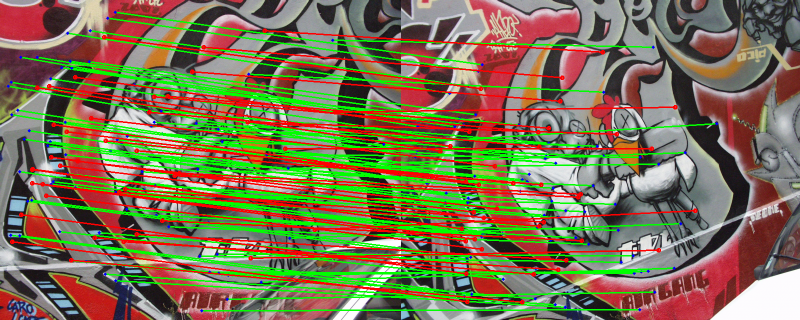}
        \caption{Graf}
        \label{fig:second_image}
    \end{subfigure}

    \vspace{-1em}
    
    \caption{Green and red lines represent inliers and outliers found by NeuroRS-CPU on two affine image registration instances.}
    \label{fig:affine_transform}
\end{figure}

We created affine registration instances from 8 scenes of VGG Dataset\footnote{https://www.robots.ox.ac.uk/~vgg/research/affine/}. In total, 40 image pairs were selected and SIFT feature matches were extracted with VLFeat toolbox~\cite{vedaldi2012vlfeat}, which were pruned using Lowe's 2nd nearest neighbor test. See Fig.~\ref{fig:affine_transform} for samples and qualitative results.

\vspace{-1em}
\paragraph{Metric} As most VGG ground-truth homographies are near-affine, the estimated affine matrix $\bH_\text{A}$ can be lifted to a full homography $\bH_{est} \in \bbR^{3\times3}$ by appending a projective row~\cite[Chap.~2]{hartley2003multiple}. We follow the evaluation protocol in~\cite{zhang2020mlifeat, sarlin2020superglue, lindenberger2023lightglue} and compute the area under the cumulative error curve (AUC). For each pair, we project the four corners using ground-truth and estimated homographies, compute the corner error, and report AUC up to a 10-pixel threshold. See, \eg,~\cite[Sec.~5.2]{sarlin2020superglue}, for details of this metric.

\vspace{-1em}
\paragraph{Competitors} We compared the performance of \neurorscpu~against RANSAC~\cite{fischler1981random} and the advanced methods: LO-RANSAC~\cite{chum2003locally}, PROSAC~\cite{chum2005matching}, Graph-cut RANSAC~\cite{barath2018graph} and MAGSAC++~\cite{barath2020magsac++}. Except RANSAC, which was implemented with numpy, other methods were based on OpenCV~\cite{bradski2000opencv}. Also, LS refinement was executed on the final consensus for all random sampling methods. We set the hyperparameters $K = 300$ for all 6 solvers, while $M = 200$ and $\alpha = 0.02$ were selected for \neurorscpu.

\vspace{-1em}
\paragraph{Results} We report the AUC at 5 and 10 pixels for these methods on two subsets: the near-affine subset (30 image pairs) and the entire dataset. As shown in Tab.~\ref{tab:affine_auc}, our \neurorscpu~achieved competitive AUCs compared to \rscpu~and OpenCV's RANSAC on both subsets. Note that the full dataset includes two non-affine scenes, which likely contribute to higher corner projection errors and lower AUC due to their projective transformations. Qualitative results from our \neurorscpu~are provided in Fig.~\ref{fig:affine_transform}.

\setlength{\tabcolsep}{3pt}
\begin{table}[h]
    \centering
    \small 
    \begin{tabular}{l|c|c|c|c}
        \hline \hline
        \toprule
        \multirow{2}{*}{\textbf{Methods}} 
        & \multicolumn{2}{c|}{Near-affine only} 
        & \multicolumn{2}{c}{Full dataset} \\
        & AUC@5 & AUC@10 & AUC@5 & AUC@10 \\
        \midrule
        RANSAC~\cite{fischler1981random} & 0.396 & 0.61 & 0.297 & 0.458 \\
        MAGSAC++~\cite{barath2020magsac++} & 0.393 & 0.599 & 0.294 & 0.449 \\
        GC-RANSAC~\cite{barath2018graph} & 0.397 & 0.61 & 0.298 & 0.458 \\
        PROSAC~\cite{chum2005matching}& 0.396 & 0.608 & 0.297 & 0.456 \\
        LO-RANSAC~\cite{chum2003locally} & 0.397 & 0.611 & 0.298 & 0.458 \\
        \neurorscpu & 0.396 & 0.608 & 0.297 & 0.456 \\
        \bottomrule
        \hline \hline
    \end{tabular}
    \caption{AUC evaluated at 5 and 10 pixels. Values closer to 1 indicate better performance.}
    \label{tab:affine_auc}
\end{table}

\section{Conclusions and future work}

We designed an SNN for robust fitting and implemented it on a real neuromorphic processor. Despite the limitations of the current neuromorphic hardware, by carefully mitigating the constraints, we were able to establish the viability of neuromorphic robust fitting and its superior energy efficiency compared to the original CPU version.

\subsection{Future work}

Our current SNN is catered to fitting linear models only. Extending to nonlinear models commonly encountered in computer vision will be useful.

The low capacity and precision of Loihi 2 prevent NeuroRF from being competitive against SOTA robust fitting methods. An interesting future work will be to implement NeuroRF on a neuromorphic cluster~\cite{intel_hala_point} and benchmark against SOTA methods.

Last but not least, building more advanced applications based on NeuroRF, such as augmented reality, visual odometry, SLAM and 3D mapping will be of interest.

\section*{Acknowledgement}
We acknowledge Intel Labs and the Intel Neuromorphic Research Community (INRC) for granting access to Loihi 2 and providing technical support. Tat-Jun Chin is SmartSat CRC Professorial Chair of Sentient Satellites.

\section*{Supplementary Material}
\renewcommand{\thesection}{\Alph{section}}
\setcounter{section}{0} 

\renewcommand{\thefigure}{AA}
\begin{figure*}[t]\centering
    \centering
    \includegraphics[width=0.8\linewidth]{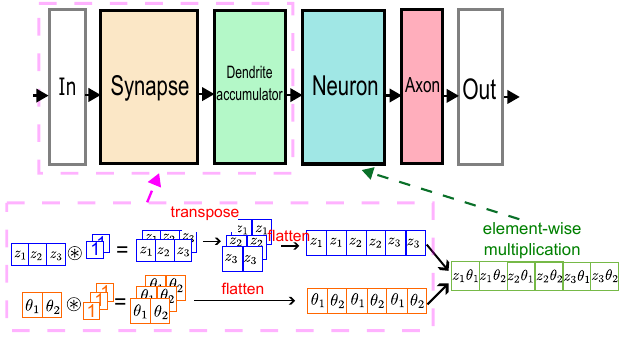}  
    \label{fig:theta_prime}
\caption{The illustration for using multiple $1\times1$ convolution filters to simulate the Kronecker product with 3 Sampling neurons $\{z_i\}_{i=1}^3$ and 2 GD neurons $\{\theta_j\}_{j=1}^2$. The $1\times1$ convolutions are defined in Synapse, while the $\operatorname{transpose(\cdot)}$ and $\operatorname{flatten(\cdot)}$ are inherently supported by Lava-Loihi framework. The element-wise multiplication is computed in Auxiliary neuron.}  
\end{figure*}

\section{Derivation for the gradient with z}
Recall the gradient descent formula for the quadratic form of least squares:
\begin{equation}
    \nabla f (\bm{\theta}) = \mathbf{Q} \bm{\theta} + \mathbf{p} 
\end{equation}
We expand
\begin{align}
    \nabla f (\bm{\theta}) & = \mathbf{Q} \bm{\theta} + \mathbf{p} \\
    & = \mathbf{Q}^\prime \left[ \begin{matrix} z_1 & 0 \\ 
    0 & z_1 \\
    z_2 & 0 \\ 
    0 & z_2 \\ 
    \vdots & \vdots \\
    z_N & 0 \\
    0 & z_N \end{matrix} \right] \bm{\theta} + \mathbf{P}^\prime \left[ \begin{matrix} z_1 \\ z_2 \\ \vdots \\ z_N \end{matrix}\right] \\
    & = \mathbf{Q}^\prime  \text{vec}\left( \left[\begin{matrix}z_1 \\ z_2 \\ \vdots \\ z_N \end{matrix} \right] \bm{\theta}^T \right) + \mathbf{P}^\prime \left[ \begin{matrix} z_1 \\ z_2 \\ \vdots \\ z_N \end{matrix}\right] \\
    & = \mathbf{Q}^\prime  \text{vec}\left( \mathbf{z}\bm{\theta}^T\right) + \mathbf{P}^\prime \mathbf{z}\\
    & = \mathbf{Q}^\prime (\mathbf{z} \otimes \bm{\theta}^T) + \mathbf{P}^{\prime} \mathbf{z}
    \label{eq:derivation_of_f_theta_z}
\end{align}
where,
\begin{equation}
    \mathbf{Q}^{\prime} = \left[ \mathbf{x}_1 \mathbf{x}_1^T \quad \mathbf{x}_2  \mathbf{x}_2^T \quad \ldots \quad \mathbf{x}_N \mathbf{x}_N^T \right] \in \mathbb{R}^{d \times Nd}
    \label{eq:Q_prime_method}
\end{equation}

\begin{equation}
    \mathbf{P}^{\prime} = \left[ -y_1 \mathbf{x_1} \quad -y_2 \mathbf{x_2} \quad \ldots \quad -y_N \mathbf{x_N} \right] \in \mathbb{R}^{d \times N}
    \label{eq:P_prime_method}
\end{equation}

Let $\bm{\theta}^{\prime} = \operatorname{vec} \left( \mathbf{z}  \bm{\theta^T}\right)$, Eq.~\eqref{eq:derivation_of_f_theta_z} becomes
\begin{equation}
    \nabla f (\bm{\theta}, \mathbf{z}) = \mathbf{Q}^{\prime} \bm{\theta}^{\prime} + \mathbf{P}^{\prime} \mathbf{z}
    \label{eq:gradient_with_z}
\end{equation}

\section{Construct NeuroRF synaptic matrices for affine transformation }

\paragraph{Problem definition} Given a point correspondence $\langle \mathbf{x}'_i, \mathbf{x}_i \rangle$, the affine transformation is defined by
\begin{equation}
    \mathbf{x}^{\prime} = \mathbf{H}_\text{A} \widetilde{\mathbf{x}}
    \label{supp:eq:aff_trans}
\end{equation}
where $\mathbf{x}^{\prime} = (x', y')$ and $\widetilde{\mathbf{x}} = [\mathbf{x}^T \quad 1] = (x, y, 1)$ is $\mathbf{x}$ in homogeneous coordinates. The $2 \times 3$ affine transformation matrix $\mathbf{H}_\text{A}$ can be estimated with 3 point correspondences~\cite[Chap 2]{hartley2003multiple}, which constitute a 6-equation linear system to solve 6 unknown parameters, i.e. $d = 6$
\begin{equation}
    \mathbf{H}_\text{A} =
    \begin{bmatrix}
        a_{11} & a_{12} & t_{x} \\
        a_{21} & a_{22} & t_{y} \\
    \end{bmatrix}
\end{equation}

Given $\mathbf{H}_\text{A}$ and Eq.~\eqref{supp:eq:aff_trans}, the residual at the point $i$  can be defined as the ``transformation error"

\begin{equation}
    r_i\left(\mathbf{H}_\text{A}\right) = \left|\mathbf{x}_i^\prime - \mathbf{H}_\text{A} \widetilde{\mathbf{x}}_i\right|
\end{equation}
RANSAC is applied to find $\mathbf{H}_\text{A}$ that maximises the consensus
\begin{equation}
    \Psi(\mathbf{H}_\text{A}) = \sum \limits_{i=1}^N \mathbb{I}\left( \left|\mathbf{x}_i^\prime - \mathbf{H}_\text{A} \widetilde{\mathbf{x}}_i\right| \leq \epsilon\right),
    \label{supp:eq:max_con_aff_trans}
\end{equation}

To apply our NeuroRF method for affine transformation, we first vectorize the affinity matrix
\begin{equation}
    \bm{\theta} = \operatorname{vec}(\mathbf{H}_{\text{A}}) \in \mathbb{R}^d
\end{equation}
where $d = 6$.

We next compute $\mathbf{Q}^{\prime}$ and $\mathbf{P}^{\prime}$ from $\mathbf{X}' \in \mathbb{R}^{N \times 2}$ and $\widetilde{\mathbf{X}} \in \mathbb{R}^{N \times 3}$ to apply the gradient descent of the ModelHypothesis (Eq.~\eqref{eq:gradient_with_z})

\begin{equation}
    \mathbf{Q}^{\prime} = 
        \begin{bmatrix}
            \widetilde{\mathbf{x}}_1 \widetilde{\mathbf{x}}_1^T & \mathbf{0}_{3 \times 3} & \ldots & \widetilde{\mathbf{x}}_N \widetilde{\mathbf{x}}_N^T & \mathbf{0}_{3 \times 3} \\
            \mathbf{0}_{3 \times 3} & \widetilde{\mathbf{x}}_1 \widetilde{\mathbf{x}}_1^T & \ldots & \mathbf{0}_{3 \times 3} & \widetilde{\mathbf{x}}_N \widetilde{\mathbf{x}}_N^T
        \end{bmatrix} \in \mathbb{R}^{d \times Nd}
    \label{eq:Q_prime_method_affine_transform}
\end{equation}

\begin{equation}
    \bm{\theta}^{\prime}  = \operatorname{vec} \left( \mathbf{z} \bm{\theta}^T \right) = \mathbf{z} \otimes \bm{\theta} \in \mathbb{R}^{Nd}
    \label{eq:theta_prime_method2}
\end{equation}

\begin{equation}
    \mathbf{P}^{\prime} = 
        \begin{bmatrix}
            -x'_1 \widetilde{\mathbf{x}}_1 & -x'_2 \widetilde{\mathbf{x}}_2 & \ldots & -x'_N \widetilde{\mathbf{x}}_N \\
            -y'_1 \widetilde{\mathbf{x}}_1 & -y'_2 \widetilde{\mathbf{x}}_2 & \ldots & -y'_N \widetilde{\mathbf{x}}_N
        \end{bmatrix} \in \mathbb{R}^{d \times N}
    \label{eq:P_prime_method_affine_transform}
\end{equation}
Note that in Eq.~\eqref{eq:P_prime_method_affine_transform}, $x'_i$ and $y'_i$ are broadcast over $\widetilde{\mathbf{x}}_i$.

\section{Runtime results for NeuroRF-CPU vs RS-CPU}
Fig.~\ref{fig:rf_on_cpu_runtime} reported the average runtime between \neurorscpu~and \rscpu~of Sec 5.2. in the Main paper. The runtime of \neurorscpu~was from a CPU simulation of our NeuroRF, hence not reflective of \neurorsloihi.

\renewcommand{\thefigure}{BB}
\begin{figure*}[t]\centering
\begin{subfigure}[b]{0.33\textwidth}
    \includegraphics[width=0.99\columnwidth]{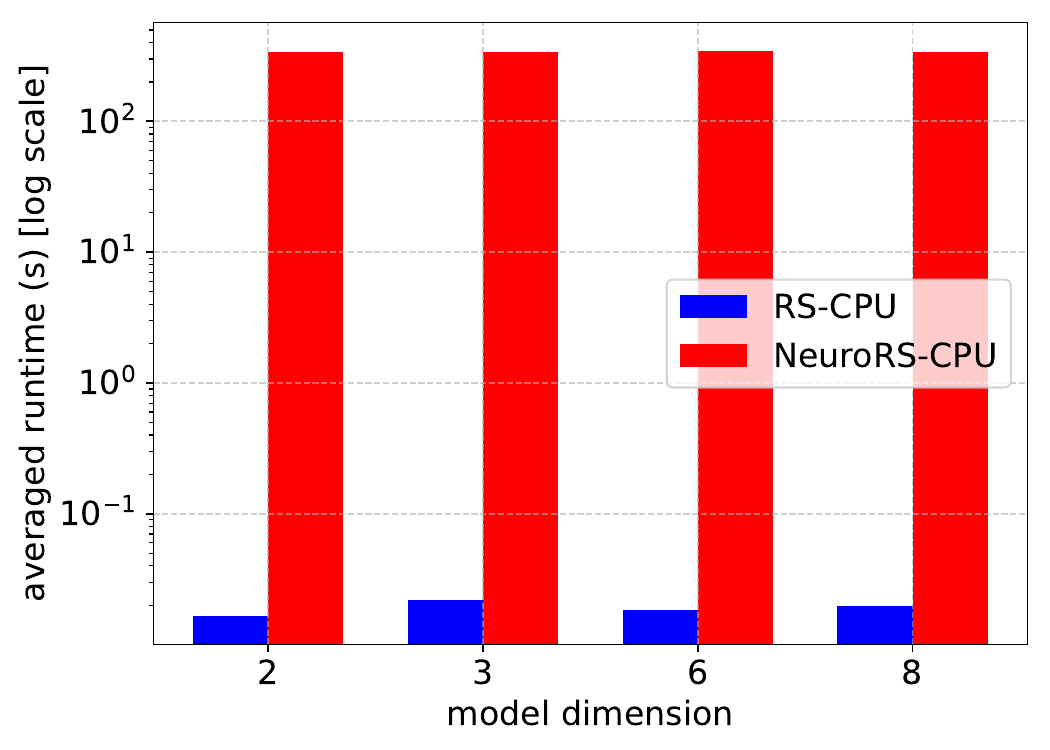}
    \caption{Vary $d$ with $(N=200, \eta = 0.2)$}
    \label{fig:rf_wrt_d_runtime}
\end{subfigure}
\begin{subfigure}[b]{0.33\textwidth}
    \includegraphics[width=0.99\columnwidth]{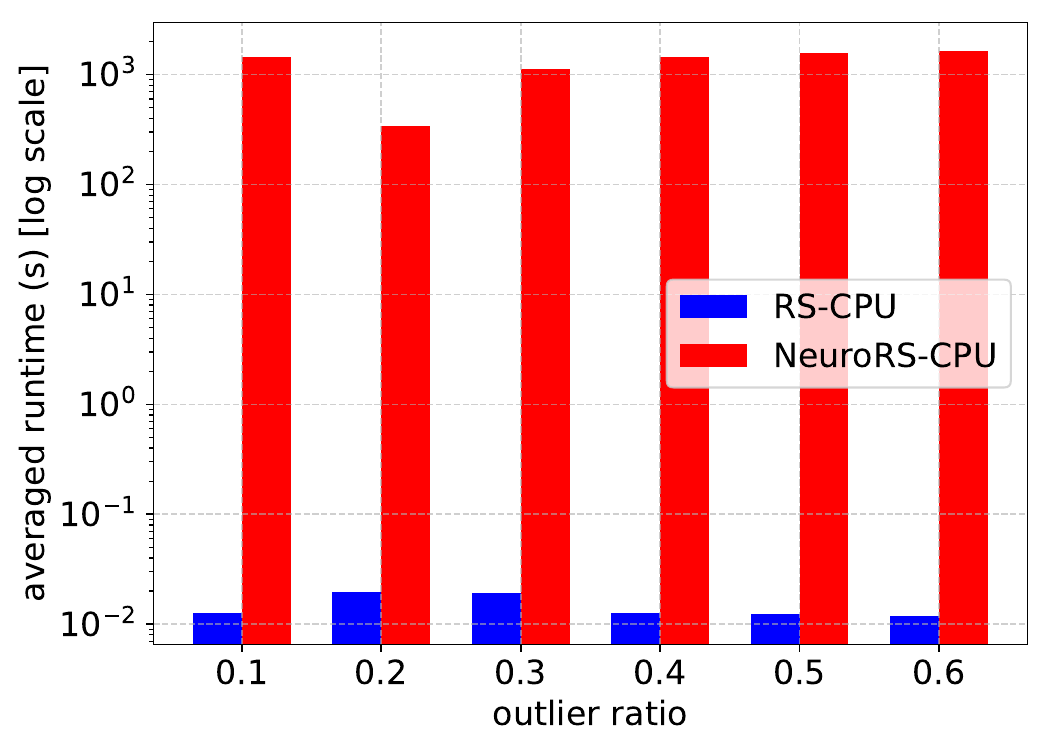}
    \caption{Effect of outlier ratio $\eta$ $(N=200, d = 8)$ }
    \label{fig:rf_wrt_eta}
\end{subfigure}
\begin{subfigure}[b]{0.33\textwidth}
    \includegraphics[width=0.99\columnwidth]{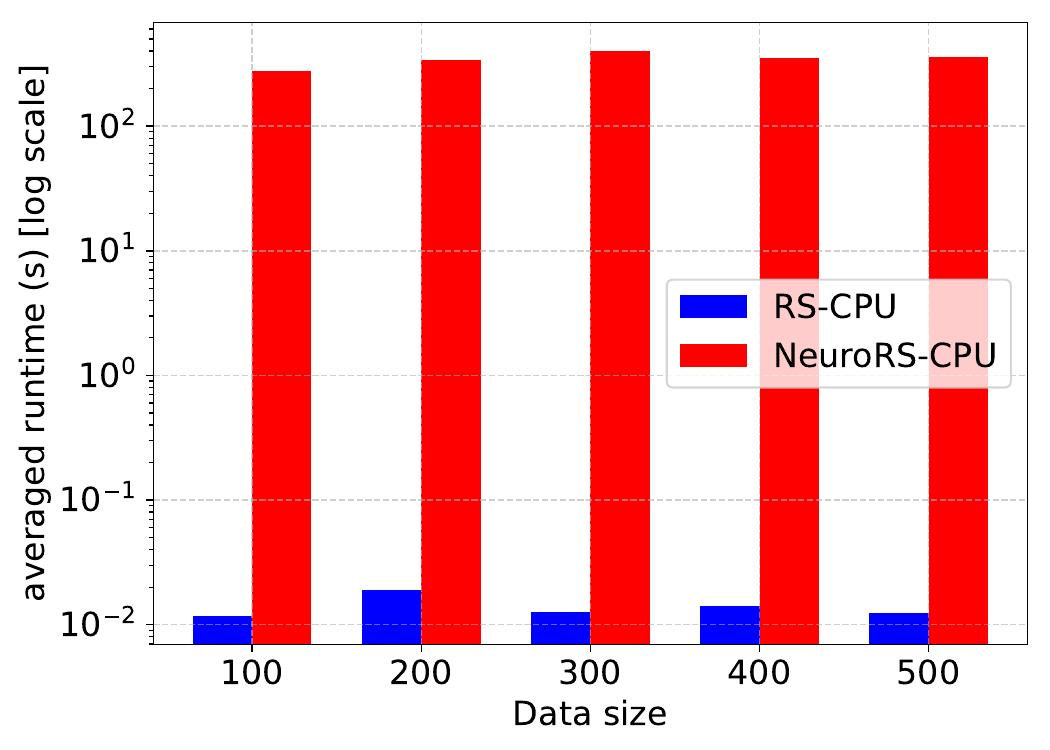}
    \caption{Vary $N$ with $(\eta = 0.2, d = 8)$}
    \label{fig:rf_wrt_N}
\end{subfigure}
\vspace{-1em}
\caption{Averaged runtime across various levels of difficulty. Results were averaged over 10 trials for each method.}
\label{fig:rf_on_cpu_runtime}
\end{figure*}

\section{Performance on neuromorphic hardware on N = 10}
We include additional results for the problem size of $N = 10, d = 2$ for Sec.~5.3 in the Main paper. Fig~\ref{fig:consensus_size_N_10} demonstrates the equivalent solution quality between \neurorsloihi~and \rscpu. Fig~\ref{fig:energy_consumption_N_10} shows that \neurorsloihi~consumes approximately $13\%$ dynamic energy compared to its competitor, which again confirms the superiority of energy efficiency of our \neurorsloihi. Fig~\ref{fig:runtime_N_10} indicates that the runtime of \neurorsloihi~was higher than \rscpu, which was probably because \neurorsloihi~solved least squares with GD, while \rscpu~computed least squares with analytical solutions.

\renewcommand{\thefigure}{CC}
\begin{figure*}[t]\centering
\begin{subfigure}[b]{0.33\textwidth}
    \includegraphics[width=0.99\columnwidth]{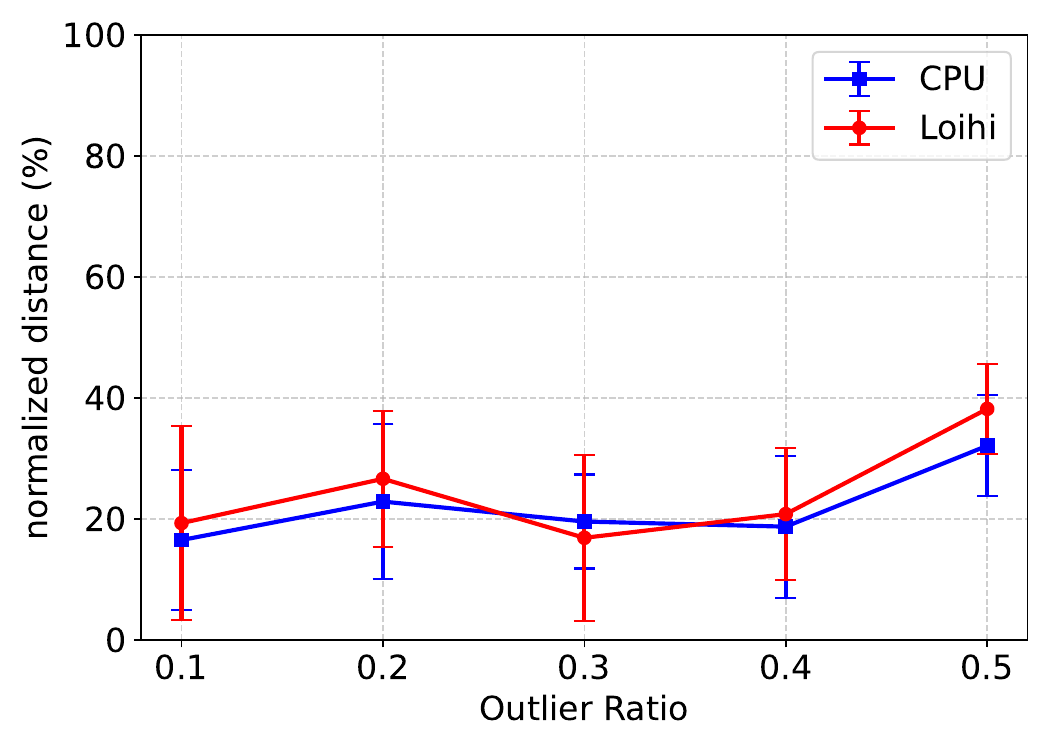}
    \caption{}
    \label{fig:consensus_size_N_10}
\end{subfigure}
\begin{subfigure}[b]{0.33\textwidth}
    \includegraphics[width=0.99\columnwidth]{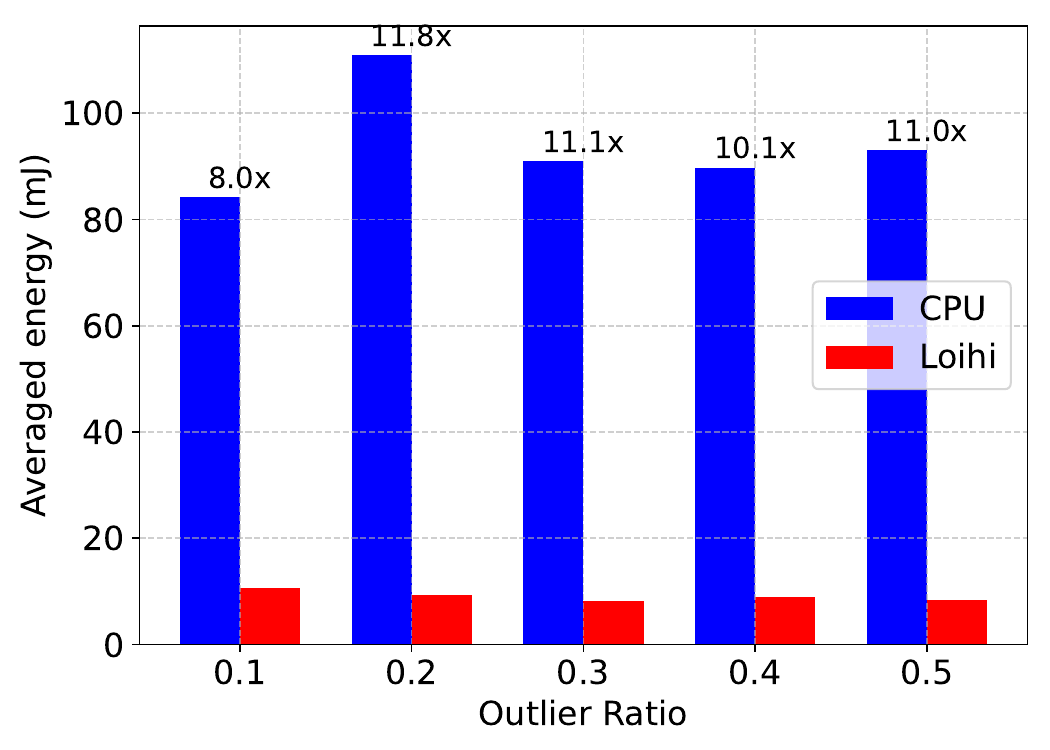}
    \caption{}
    \label{fig:energy_consumption_N_10}
\end{subfigure}
\begin{subfigure}[b]{0.33\textwidth}
    \includegraphics[width=0.99\columnwidth]{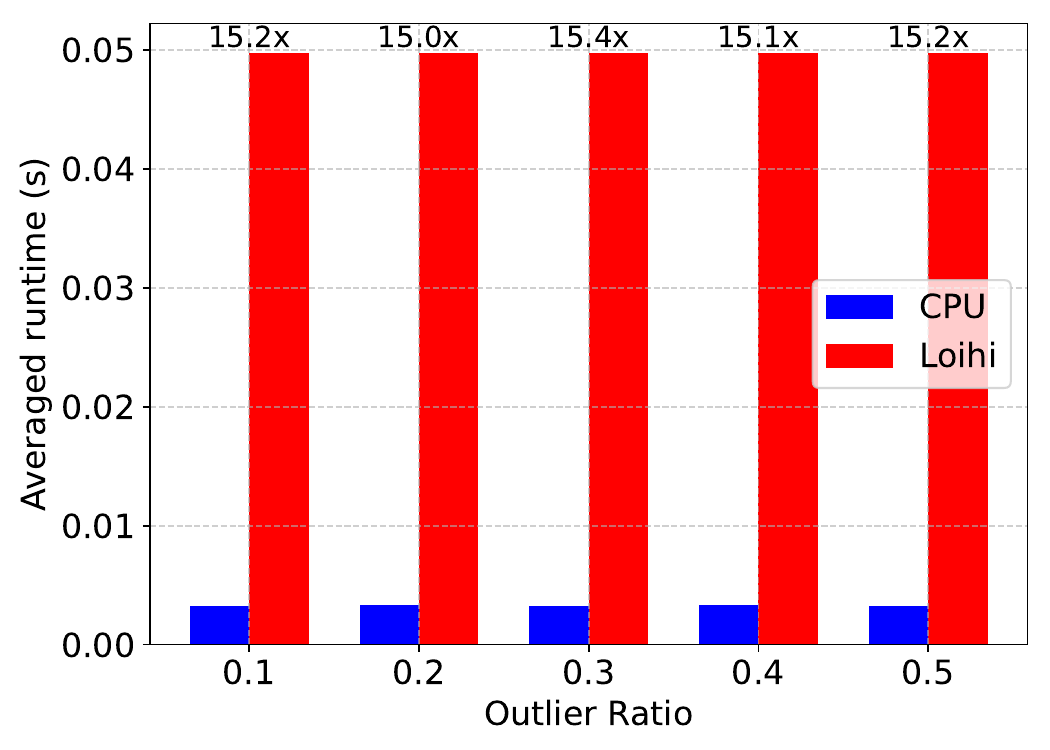}
    \caption{}
    \label{fig:runtime_N_10}
\end{subfigure}
\vspace{-1em}
\caption{Performance on neuromorphic hardware. (a) Average consensus size, (b) average dynamic energy consumption and (c) average runtime of \neurorsloihi~and \rscpu~on synthetic line fitting instances with $N = 10$ points, plotted against outlier rate.}
\label{fig:hardwareperformance_N_10}
\end{figure*}

{
    \small
    \bibliographystyle{ieeenat_fullname}
    \bibliography{main}
}

\end{document}